\documentclass[twocolumn]{article}

\usepackage{PRIMEarxiv}

\usepackage[utf8]{inputenc} 
\usepackage[T1]{fontenc}    
\usepackage{hyperref}       
\usepackage{url}            
\usepackage{booktabs}       
\usepackage{amsfonts}       
\usepackage{nicefrac}       
\usepackage{microtype}      
\usepackage{fancyhdr}       
\usepackage{graphicx}       
\graphicspath{{media/}}     
\usepackage{xcolor}
\usepackage{cite}
\usepackage{amsmath,amssymb}
\usepackage{algorithmic}
\usepackage{algorithm}
\usepackage{textcomp}
\usepackage{enumitem}
\usepackage{marvosym}
\setlist[itemize]{leftmargin=1em,itemsep=0pt,parsep=0pt,topsep=3pt}

\title{\raisebox{-0.5\height}{\includegraphics[height=1.2em]{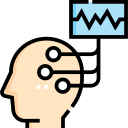}}~EVA-Net: Interpretable Anomaly Detection for Brain Health via Learning Continuous Aging Prototypes from One-Class EEG Cohorts}

\author{
  \begin{tabular}[t]{c@{\hskip 3em}c}
  \begin{tabular}[t]{c}
  Kunyu Zhang \\
  Qilu Hospital, ASU \\
  \texttt{kzhan195@asu.edu}
  \end{tabular}
  &
  \begin{tabular}[t]{c}
  Mingxuan Wang \\
  Zhengzhou University 
  \end{tabular}
  \\[3em]
  \begin{tabular}[t]{c}
  Xiangjie Shi \\
  University of Science and \\
  Technology Beijing
  \end{tabular}
  &
  \begin{tabular}[t]{c}
  Haoxing Xu \\
  Southern University of \\
  Science and Technology
  \end{tabular}
  \\[3em]
  \multicolumn{2}{c}{
  \begin{tabular}[t]{c}
  Chao Zhang\textsuperscript{\Letter} \\
  Qilu Hospital \\
  \texttt{chao\_zhang@sdu.edu.cn}
  \end{tabular}
  }
  \end{tabular}
}

\begin{document}

\twocolumn[
\begin{@twocolumnfalse}
\maketitle

\begin{abstract}
The brain age is a key indicator of brain health. While electroencephalography (EEG) is a practical tool for this task, existing models struggle with the common challenge of imperfect medical data, such as learning a ``normal'' baseline from weakly supervised, healthy-only cohorts. This is a critical anomaly detection task for identifying disease, but standard models are often black boxes lacking an interpretable structure. We propose EVA-Net, a novel framework that recasts brain age as an interpretable anomaly detection problem. EVA-Net uses an efficient, sparsified-attention Transformer to model long EEG sequences. To handle noise and variability in imperfect data, it employs a Variational Information Bottleneck to learn a robust, compressed representation. For interpretability, this representation is aligned to a continuous prototype network that explicitly learns the normative healthy aging manifold. Trained on 1297 healthy subjects, EVA-Net achieves state-of-the-art accuracy. We validated its anomaly detection capabilities on an unseen cohort of 27 MCI and AD patients. This pathological group showed significantly higher brain-age gaps and a novel Prototype Alignment Error, confirming their deviation from the healthy manifold. EVA-Net provides an interpretable framework for healthcare intelligence using imperfect medical data.
\end{abstract}

\vspace{0.5cm}
\end{@twocolumnfalse}
]

\section{Introduction}
\label{sec:intro}

Brain age (BA) aims to estimate an individual's biological age of the brain from neural data, and the difference between the estimated BA and chronological age, known as the brain-age gap (BAG), serves as a compact indicator of brain health and aging velocity~\cite{peng2021accurate,lee2022deep,de2022mind}. Compared with structural or functional neuroimaging pipelines, electroencephalography (EEG) offers practical advantages in cost, accessibility, wearability, and longitudinal monitoring, while directly capturing fast neural dynamics~\cite{siuly2016eeg,smith2005eeg,ros2013mind}. These properties position EEG-based BA modeling as a complementary route to imaging-based approaches for population-level health assessment and individual risk stratification, and make it well-suited for tracking functional changes related to aging, lifestyle, and interventions.

Despite promising progress, current EEG BA studies face several methodological bottlenecks~\cite{roy2019deep}. First, efficient long-range temporal modeling remains limited: many methods rely on short windows or local convolutions and therefore struggle to capture multi-scale dynamics spanning instantaneous waveforms to segment-level rhythms under tractable computation. Second, robustness and cross-domain generalization are challenged by device, site, and subject variability~\cite{liu2025advancing}. Standard models trained on single-source data often overfit to site-specific noise and artifacts~\cite{zhang2025clinical}, which impairs their transferability to external clinics. EEG is sensitive to these distribution shifts, challenging stability on unseen data. Third, and most critically for real-world application, standard methods fail to address the challenge of imperfect medical data, such as learning from one-class (e.g., healthy-only) cohorts. This is a form of anomaly detection for disease detection using imperfect data, where the goal is not just regression, but to define a tight, interpretable boundary of ``normal'' aging. Current end-to-end models lack an explicit ``ideal healthy aging trajectory,'' making it difficult to perform principled anomaly detection or interpret the BAG as a true deviation from health, rather than just statistical noise.

To address these issues, we propose a unified framework that recasts brain age prediction as a weakly supervised anomaly detection task~\cite{li2025age,gao2025self}, designed to operate on imperfect (healthy-only) medical data. Concretely, short-window EEG epochs are first transformed via temporal embedding into a sequence representation amenable to efficient attention~\cite{liu2024eeg2video}. A sparsified long-sequence attention backbone then captures multi-scale dependencies without sacrificing temporal coverage. In the latent space, a variational information bottleneck (VIB) enhances robustness by suppressing age-irrelevant factors linked to subject- or device-specific variability. This acts as a regularization mechanism that forces the model to `forget' site-specific nuisance variables from the training source, thereby preserving intrinsic aging patterns that are transferable to unseen domains. We further introduce an age-conditioned continuous prototype network that maps scalar age onto a smooth, normative healthy trajectory in latent space~\cite{zhang2024improving}. An alignment constraint pulls healthy sample embeddings toward their age-conditioned prototypes, endowing the representation with an explicit geometrical structure for ``normal'' age. This transforms the model into a powerful anomaly detector~\cite{deng2021graph}, where pathological states (e.g., AD) can be identified and quantified as significant deviations from this interpretable healthy manifold.

The main contributions of this work are as follows:
\begin{itemize}
    \item \textbf{Efficient long-sequence temporal modeling.} We employ a sparsified long-sequence attention encoder as the backbone to unify multi-scale dependencies from instantaneous waveforms to segment-level rhythms, offering a large receptive field under tractable computation and avoiding the fragmentation of short-window processing.
    \item \textbf{Robust representation via a variational information bottleneck.} We impose an explicit information-throughput constraint in the latent space to suppress age-irrelevant subject/device factors and noise. This enforces a selective compression of healthy age-related features, yielding a robust representation of normative aging and ensuring the learned manifold is not contaminated by non-age-related variance when learning from imperfect (healthy-only) medical data.
    \item \textbf{Age-conditioned continuous prototype alignment.} We construct a continuous ``normative healthy aging trajectory'' by mapping scalar age into the latent space and align sample embeddings to their age-conditioned prototypes. This enforces a smooth, interpretable geometry for healthy aging, transforming the task into a weakly supervised anomaly detection problem where the BAG serves as a principled measure of deviation from the normative manifold.
    \item \textbf{Validation as an Anomaly Detector.} Across strong baselines and diverse data settings, our method achieves \emph{overall superior} performance in accuracy and robustness. We further validate our model's utility as an anomaly detector by demonstrating that an unseen cohort of AD patients shows a significantly larger BAG, confirming their deviation from the learned healthy aging manifold.
\end{itemize}

\section{Related Work}
\label{sec:related_work}

\subsection{Deep learning for brain age prediction}
Brain age (BA) and the brain-age gap (BAG) have been widely explored using deep learning in neuroimaging~\cite{peng2021accurate,lee2022deep,beheshti2021predicting}. CNNs capture morphometric patterns, graph models encode region interactions, and Transformers aggregate multi-scale context~\cite{vaswani2017attention}. Recent advances include multimodal imaging fusion~\cite{kassani2019multimodal} and disease-specific models~\cite{abrol2025generating,wei2025symptomatic}. However, most rely on mixed cohorts or case-control designs. Few address learning normative trajectories from healthy-only cohorts, essential for defining a baseline to detect pathological deviations~\cite{de2022mind}. Imaging approaches also face cost and accessibility constraints. Our work leverages efficient long-sequence attention for EEG, designed for normative modeling and anomaly detection.

\subsection{Development of EEG-based brain age prediction}
EEG-based BA research combines handcrafted features with regressors, end-to-end learning via CNNs and spatiotemporal graphs~\cite{wang2025bigdc,shan2022spatial}, and robustness techniques including domain adaptation~\cite{wu2024multi}. Transformers have captured long-range dependencies~\cite{lee2022eeg,wang2024eegpt,gemein2024brain}. Three challenges remain for anomaly detection from imperfect data: (i) efficient long-range temporal modeling; (ii) resilience to distribution shifts; and (iii) lack of latent geometry reflecting normative aging, preventing BAG interpretation as an anomaly score. We address these with a sparsified encoder (i), variational information bottleneck (ii), and age-conditioned prototype network (iii).

\subsection{Explainability and latent-space structure}
Explainability for BA models includes saliency visualizations, prototype-based approaches~\cite{nauta2023pip,wang2025protomol}, and latent-space regularization. The Information Bottleneck (IB)~\cite{tishby2015deep,hu2024survey} learns compressed representations retaining task-relevant information and has been applied to brain disorder diagnosis~\cite{zhang2025mvho}. The Variational Information Bottleneck (VIB)~\cite{alemi2016deep} extends this via tractable variational formulation for weakly-supervised tasks~\cite{li2023task}. However, most strategies remain post hoc and lack continuous healthy age manifolds for BAG interpretation. Recent normative modeling~\cite{attye2024data} explores generative manifold learning. Our framework integrates explanation into design: VIB enforces compression, and age-conditioned prototypes shape latent space into a smooth healthy aging trajectory, providing interpretable anomaly detection.

\section{Methodology}
\label{sec:method}
\begin{figure*}
    \centering
    \includegraphics[width=0.96\linewidth]{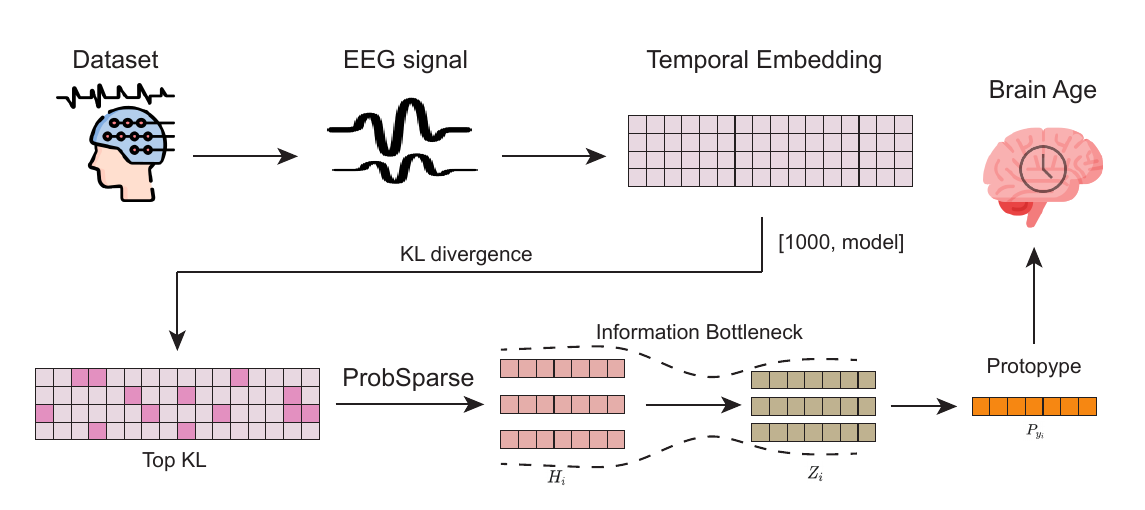}
    \caption{The overall architecture of our proposed EVA-Net. EEG signals are converted via Temporal Embedding into a sequence. An efficient backbone (using ProbSparse attention by identifying Top KL time-points) extracts a hidden state $H_i$. This state is passed through an Information Bottleneck (VIB) to produce a robust latent code $Z_i$. $Z_i$ is then used for both Brain Age prediction (regression) and alignment with an ideal age-matched Prototype $P_{y_i}$.}
    \label{fig:framework}
\end{figure*}

In this section, we present \textbf{EVA-Net}, a novel deep learning Efficient Variational Alignment Network designed to learn a robust and interpretable normative manifold of healthy brain aging, as shown in Figure~\ref{fig:framework}. The overall architecture comprises three key components: (1) an \textbf{E}fficient long-sequence encoder with ProbSparse attention to capture multi-scale temporal dependencies; (2) a \textbf{V}ariational Information Bottleneck (VIB) to learn a compressed and robust representation $Z_i$ by suppressing age-irrelevant noise; and (3) an \textbf{A}lignment module with an age-conditioned continuous prototype network ($P_y$) that provides an interpretable structure for healthy aging. By jointly optimizing for prediction accuracy, representation compression ($L_{IB}$), and prototype alignment ($L_{align}$), EVA-Net enables the robust and interpretable detection of anomalies as deviations from the learned healthy manifold.

\subsection{Problem Formulation}
\label{sec:problem_formulation}
Our objective is to learn a normative model of healthy brain aging from a training set $D_{train} = \{(X_i, y_i)\}_{i=1}^{N}$ of $N$ healthy subjects. The input $X_i \in \mathbb{R}^{C \times T}$ is a 19-channel, 4-second (T=1000) EEG epoch, and $y_i \in \mathbb{R}^+$ is the chronological age. Our framework, EVA-Net, is trained to learn three functions simultaneously: (1) an efficient probabilistic encoder $E_{\phi}$ (Informer-VIB backbone) that maps $X_i$ to a robust latent representation $Z_i \in \mathbb{R}^d$ by optimizing for efficiency (ProbSparse) and robustness (VIB); (2) a continuous prototype network $P_{\theta}$ that maps a true scalar age $y$ to its corresponding ideal prototype $P_y \in \mathbb{R}^d$ in the same latent space; and (3) a prediction head $f_{\psi}$ that regresses age $\hat{y}_i = f_{\psi}(Z_i)$. All parameters $\{\phi, \theta, \psi\}$ are jointly optimized via a composite loss $L_{Total} = L_{pred} + \beta L_{IB} + \gamma L_{align}$ to balance prediction accuracy, information compression, and geometric alignment. For anomaly detection, this allows us to compute both the standard Brain-Age Gap (BAG, $\hat{y} - y$) and a novel Prototype Alignment Error (PAE, $\| Z - P_y \|_2$) as an interpretable measure of deviation from the learned healthy manifold.

\subsection{Efficient Long-Sequence Encoder}
\label{sec:efficient_encoder}
\begin{figure}
    \centering
    \includegraphics[width=0.96\linewidth]{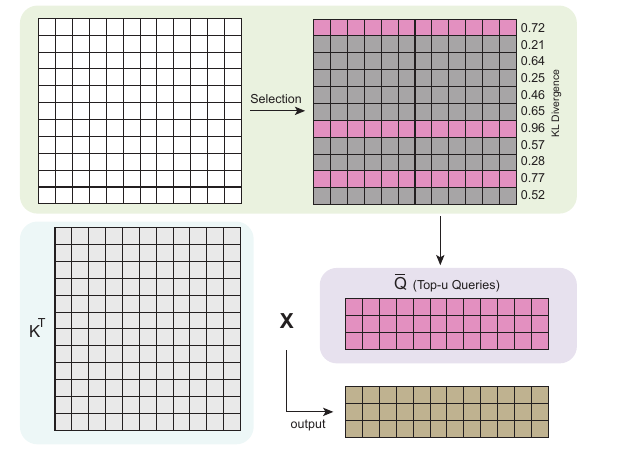}
    \caption{ProbSparse attention mechanism achieving $O(L \log L)$ complexity. Queries are ranked by KL divergence scores, with top-scoring queries (pink) forming sparse matrix $\bar{Q}$. Computing $\bar{Q}K^T$ bypasses full $L \times L$ attention, reducing computational cost while preserving critical temporal dependencies.}
    \label{fig:placeholder}
\end{figure}

The first component processes input $X_i \in \mathbb{R}^{C \times T}$ ($C=19$ channels, $T=1000$ time-points). Standard Transformer self-attention
\begin{equation}
    \mathcal{A}(Q, K, V) = \text{Softmax}(\frac{QK^T}{\sqrt{d_k}})V,
\end{equation}
requires $O(T^2)$ computation, prohibitive for $T=1000$. Channel-wise features are projected to $d_{model}$ with sinusoidal positional encoding, yielding $S_i \in \mathbb{R}^{T \times d_{model}}$.

The \textbf{ProbSparse} attention reduces complexity to $O(T \log T)$ (Figure~\ref{fig:placeholder}) by selecting queries with high KL divergence from uniform attention. Query importance is approximated by:
\begin{equation}
    \bar{M}(q_i, K) = \max_{j} \left\{ \frac{q_i k_j^T}{\sqrt{d_k}} \right\} - \frac{1}{T} \sum_{j=1}^{T} \frac{q_i k_j^T}{\sqrt{d_k}}
\end{equation}
We select the top $u = c \cdot \ln T$ queries to form sparse set $\hat{Q}$, computing attention only for this subset:
\begin{equation}
    \text{A}_{ProbSparse}(Q, K, V) = \text{Softmax}\left(\frac{\hat{Q}K^T}{\sqrt{d_k}}\right)V
\end{equation}
After $L$ encoder layers, global average pooling produces hidden representation $H_i \in \mathbb{R}^{d_{model}}$.

\subsection{Variational Information Bottleneck}
\label{sec:vib}
The hidden representation $H_i \in \mathbb{R}^{d_{model}}$ produced by the efficient encoder, while compact, may still contain substantial age-irrelevant information, such as subject-specific, site-specific, or noise-related features inherent in imperfect medical data. To enhance robustness and generalization, we introduce a Variational Information Bottleneck (VIB) as a probabilistic regularizer. This module ensures that the final latent representation $Z_i$ is selectively compressed to retain only the minimal information necessary for age prediction.

Instead of a deterministic mapping, we treat $H_i$ as the input to a stochastic encoder that parameterizes a posterior distribution $q_{\phi}(Z_i | X_i)$, which we approximate as a Gaussian $\mathcal{N}(\mu_i, \sigma_i^2)$. The mean $\mu_i \in \mathbb{R}^d$ and log-variance $\log\sigma_i^2 \in \mathbb{R}^d$ (where $d$ is the latent dimension) are computed via two separate linear heads applied to $H_i$:
\begin{equation}
    \mu_i = W_{\mu}H_i + b_{\mu} \quad \text{and} \quad \log\sigma_i^2 = W_{\sigma}H_i + b_{\sigma}
\end{equation}
To enable backpropagation, we use the reparameterization trick to sample the final latent vector $Z_i \in \mathbb{R}^d$:
\begin{equation}
    Z_i = \mu_i + \sigma_i \odot \epsilon, \quad \text{where } \epsilon \sim \mathcal{N}(0, I)
\end{equation}

This stochastic encoding introduces a new objective to the total loss function: the Information Bottleneck loss, $L_{IB}$. This loss is the KL divergence between the learned posterior $q_{\phi}(Z_i | X_i)$ and a standard Gaussian prior $p(Z) = \mathcal{N}(0, I)$. This $L_{IB}$ term, defined as:
\begin{equation}
    L_{IB} = D_{KL}(q_{\phi}(Z_i | X_i) || p(Z)) = \frac{1}{2} \sum_{j=1}^{d} (\sigma_{i,j}^2 + \mu_{i,j}^2 - \log\sigma_{i,j}^2 - 1)
    \label{eq:ib}
\end{equation}
acts as a powerful regularizer. It penalizes the information throughput, forcing the encoder to discard the non-essential variations in $X_i$ (noise, artifacts, subject ID) and produce a maximally compressed, robust representation $Z_i$ focused solely on healthy aging.

\subsection{Continuous Prototype Alignment}
\label{sec:prototype_alignment}

The VIB module ensures the latent representation $Z_i$ is robust and compressed. However, it does not guarantee that the latent space is well-structured; the representations of subjects with similar ages (e.g., 60 and 61) are not explicitly encouraged to be close to each other. To address this and endow the model with a clear, interpretable geometry, we introduce the Alignment component: an age-conditioned continuous prototype network, $P_{\theta}$, parameterized by $\theta$.

This independent network $P_{\theta}$ is designed to learn the ``ideal'' manifold of healthy aging by mapping a scalar chronological age $y \in \mathbb{R}^+$ to its corresponding ``ideal'' prototype $P_y \in \mathbb{R}^d$ in the same latent space as $Z_i$. A simple scalar age is a poor input for a neural network; therefore, we first encode $y_i$ into a high-dimensional vector $\mathbf{y}_i^{embed}$ using a sinusoidal (Fourier) positional encoding, analogous to that used in Transformers. This age embedding is then passed through $P_{\theta}$, which is implemented as a small Multi-Layer Perceptron (MLP):
\begin{equation}
    P_{y_i} = P_{\theta}(\mathbf{y}_i^{embed})
\end{equation}
To enforce the geometric structure, we introduce a loss term, the prototype alignment loss $L_{align}$. This loss acts as a ``pulling'' force, compelling the probabilistic encoder $E_{\phi}$ to map its sample representation $Z_i$ to be close to the ideal prototype $P_{y_i}$ defined by the prototype network. We define this loss as the Mean Squared Error (MSE) between the two vectors:
\begin{equation}
    L_{align} = \| Z_i - P_{y_i} \|_2^2
    \label{eq:align}
\end{equation}
This alignment loss is critical, as it forces the encoder $E_{\phi}$ and the prototype network $P_{\theta}$ to jointly discover a latent space where the ``ideal'' healthy aging trajectory $P_y$ is continuous and monotonic, and where the ``real'' (and noisy) sample embeddings $Z_i$ are anchored to their respective positions along this normative manifold.

\subsection{Training Objective and Composite Loss}
\label{sec:training_objective}

The EVA-Net framework is trained end-to-end by jointly optimizing all three components: the efficient encoder $E_{\phi}$, the prototype network $P_{\theta}$, and a final prediction head $f_{\psi}$. The prediction head is a simple Multi-Layer Perceptron (MLP) that takes the robust latent vector $Z_i \in \mathbb{R}^d$ (from the VIB module) as input and regresses the final scalar brain age $\hat{y}_i$:
\begin{equation}
    \hat{y}_i = f_{\psi}(Z_i)
\end{equation}
Our full training objective is a composite loss $L_{Total}$ designed to simultaneously balance three distinct goals: prediction accuracy, representation robustness, and latent space interpretability. This loss is the weighted sum of the three corresponding loss terms:
\begin{equation}
    L_{Total} = L_{pred} + \beta \cdot L_{IB} + \gamma \cdot L_{align}
\end{equation}
where $\beta$ and $\gamma$ are hyperparameters that balance the contribution of each component.

The first term, $L_{pred}$, is the primary task objective, ensuring the latent representation $Z_i$ contains sufficient information to accurately predict age. We define this as the Mean Squared Error (MSE) between the predicted age and the true chronological age:
\begin{equation}
    L_{pred} = \frac{1}{N} \sum_{i=1}^{N} (\hat{y}_i - y_i)^2
\end{equation}
The second term, $L_{IB}$, is the KL divergence loss from the VIB module (defined in Eq.~\ref{eq:ib}). This term acts as an information-theoretic regularizer, penalizing the model for encoding age-irrelevant information and thus enhancing the robustness and generalization of $Z_i$.
The third term, $L_{align}$, is the prototype alignment loss (defined in Eq.~\ref{eq:align} ). This loss compels the encoder to map samples onto a structured, continuous manifold that aligns with the ``ideal'' healthy aging trajectory learned by the prototype network $P_{\theta}$. By jointly optimizing this composite loss, EVA-Net learns a latent space that is not only predictive and robust, but also geometrically interpretable.

\section{Experiments}
\label{sec:exp}
\begin{figure}
    \centering
    \includegraphics[width=0.98\linewidth]{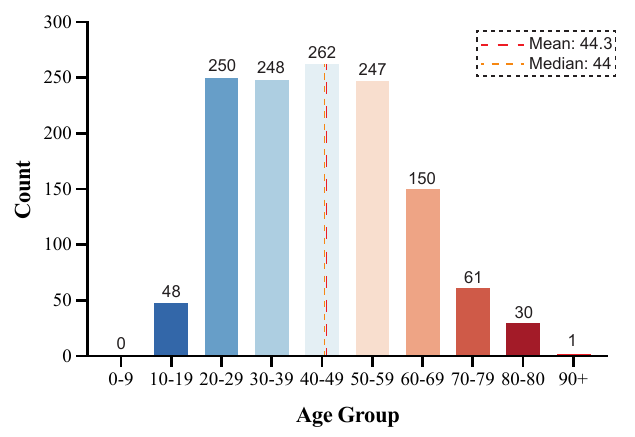}
    \caption{Age distribution of the normative healthy cohort. A bar chart illustrating the sample count by decade. The distribution is concentrated in adulthood, and fewer subjects in the younger and older brackets, with a cohort mean age of 44.3 years and a median age of 44 years.}
    \label{fig:age_distribution}
\end{figure}
\subsection{Datasets}
This study was conducted in accordance with the Declaration of Helsinki and was approved by the Ethics Committee of Qilu Hospital of Shandong University (Protocol Number: KYLL-202406 (YJ) -010). Written informed consent was obtained from all subjects or their legal guardians involved in the study.

All data were acquired from Qilu Hospital of Shandong University and divided into two distinct cohorts to train and evaluate our normative modeling framework. The primary ``normative'' cohort consists of 1297 healthy subjects selected based on clinical reports documenting a ``normal'' EEG and the absence of any neurological or psychiatric disorders. As shown in Figure~\ref{fig:age_distribution}, this cohort has a mean chronological age of 44.3 years and a median of 44 years. The age distribution is not uniform; it is heavily concentrated in adulthood (20-59 years), with fewer subjects in the pediatric (10-19) and elderly (60+) age brackets. The second ``pathological'' cohort, used for out-of-distribution anomaly detection, comprises 27 subjects with cognitive impairment: 12 with Mild Cognitive Impairment (MCI) and 15 with Alzheimer's Disease (AD). 

All recordings from both cohorts were processed through an identical, standardized preprocessing pipeline (detailed in Section ~\ref{sec:dp}) to produce 19-channel, 4-second EEG epochs ($X_i \in \mathbb{R}^{19 \times 1000}$). To robustly evaluate model performance on the 1297-subject healthy cohort, we employed a 10-fold cross-validation strategy. The cohort was randomly partitioned into 10 equal, non-overlapping folds. In each of the 10 iterations (folds), one fold was reserved as the healthy test set, and the remaining nine folds were used as the training set to learn the EVA-Net parameters $\{\phi, \theta, \psi\}$. The final regression performance on healthy data is reported as the average and standard deviation of the metrics across all 10 folds. The 27 MCI and AD subjects are used as a separate test set to evaluate the model's ability to detect pathological deviations from the healthy norm.

\begin{table*}[!ht]
    \centering
    \caption{Performance comparison on the held-out healthy test set. Results are averaged across the 10-fold cross-validation. For MAE and RMSE, lower is better ($\downarrow$). For $R^2$, higher is better ($\uparrow$). Best results are highlighted in bold.}
    \label{tab:performance_comparison}
    \begin{tabular*}{\textwidth}{@{\extracolsep{\fill}}lccccccc}
        \toprule
        \textbf{Metric} & \textbf{EEGNet} & \textbf{TCN} & \textbf{Deep-EMD} & \textbf{WST} & \textbf{NCV} & \textbf{MBN} & \textbf{EVA-Net (Ours)} \\
        \midrule
        MAE $\downarrow$    & 6.435  & 4.924 & 3.497    & 3.176 & 5.172 & 3.053 & \textbf{2.645}   \\
        RMSE $\downarrow$   & 9.267  & 6.137 & 4.196    & 4.469 & 6.724 & 3.664 & \textbf{3.327}   \\
        $R^2$ $\uparrow$    & 0.836  & 0.891 & 0.933    & 0.935 & 0.857 & 0.936 & \textbf{0.939}   \\
        \bottomrule
    \end{tabular*}
\end{table*}

\subsection{Data Preprocessing}
\label{sec:dp}
All data from both the healthy and pathological cohorts underwent an identical, standardized preprocessing pipeline implemented using MNE-Python. First, raw data was loaded from CSV, units were converted to Volts, and only the 19 standard 10-20 EEG channels were retained. Bad channels were automatically detected based on abnormal variance and low inter-channel correlation criteria and subsequently repaired using spherical spline interpolation. The data was then re-referenced to a common average reference (AR) and temporally filtered using a 0.5-45 Hz FIR band-pass filter and a 60 Hz notch filter to remove signal drifts and powerline noise. Ocular, muscular, and cardiac artifacts were automatically identified and removed using FastICA, with artifactual components selected based on their high correlation (threshold: 0.3) with frontal (FP1, FP2) channels. Finally, the cleaned, continuous signal was epoched into 4-second non-overlapping windows. Any epochs with a peak-to-peak amplitude exceeding a 150 µV threshold were automatically discarded, and the resulting clean epochs (tensors of shape `[19, 1000]`) were stored and used for all subsequent model training and evaluation.

\subsection{Baselines}
We benchmark the performance of our proposed EVA-Net against a comprehensive suite of competitive baselines, all trained and evaluated on the same preprocessed datasets and 10-fold cross-validation splits. These baselines are categorized into three groups: (1) two widely-used time series models, EEGNet~\cite{eegnet} and a Temporal Convolutional Network (TCN)~\cite{TCN}; (2) two state-of-the-art (SOTA) EEG representation models, DEEP-EMD~\cite{deep-emd} and WST~\cite{wst}; and (3) two recent models specifically designed for brain age prediction, NCV~\cite{ncv} and MBN~\cite{mbn}. All baseline models were re-implemented, and their experiments were conducted following the settings described in their respective original publications.

\subsection{Implementation Details}
Our EVA-Net framework was implemented in PyTorch. All experiments are conducted on an NVDIA RTX 4090 GPU. The efficient encoder backbone was configured with $L=4$ Informer layers, a model dimension of $d_{model}=128$, and $H=8$ attention heads. The ProbSparse attention sampling factor $c$ was set to 5. The VIB module projected the $d_{model}=128$ hidden state $H_i$ into a latent dimension of $d=64$, producing $\mu_i$ and $\log\sigma_i^2$. The continuous prototype network $P_{\theta}$ and the prediction head $f_{\psi}$ were both implemented as 3-layer MLPs with ReLU activations, with $P_{\theta}$ mapping the encoded age to $d=64$ and $f_{\psi}$ mapping $Z_i$ to a single scalar output. All models were trained for 200 epochs using the AdamW optimizer with a learning rate of $1e-4$, a weight decay of $1e-5$, and a batch size of 64. A cosine annealing scheduler was used to adjust the learning rate. We applied early stopping with a patience of 20 epochs based on the validation set's $L_{pred}$ loss. The loss hyperparameters were set to $\beta=1e-3$ for the $L_{IB}$ loss and $\gamma=0.7$ for the $L_{align}$ loss based on empirical tuning on the validation set.

\subsection{Evaluation Metrics}
Our evaluation protocol is twofold, designed to assess both standard regression accuracy on healthy individuals and the model's capability for anomaly detection. 
\textbf{(1) Normative Regression Performance:} We evaluate the model's prediction accuracy on the healthy test folds from our 10-fold cross-validation. For this ``healthy person prediction standard,'' we report the average and standard deviation of three core regression metrics: the Mean Absolute Error (MAE), the Root Mean Squared Error (RMSE), and the R-squared ($R^2$) score to measure the proportion of variance explained by the model's predictions.
\textbf{(2) Anomaly Detection Efficacy:} We then evaluate the model's ability to distinguish the pathological cohort (12 MCI and 15 AD subjects) from the healthy test subjects. For this, we formally define the two metrics introduced in our Problem Formulation. The first is the standard Brain-Age Gap (BAG), calculated as the predicted age $\hat{y}$ (from $f_{\psi}$) minus the chronological age $y$:
\begin{equation}
    \text{BAG} = \hat{y} - y
\end{equation}
The second is our proposed Prototype Alignment Error (PAE), which quantifies an individual's deviation from the learned healthy manifold as the Euclidean distance between their latent embedding $Z$ (from $E_{\phi}$) and their age-matched ideal prototype $P_y$ (from $P_{\theta}$):
\begin{equation}
    \text{PAE} = \| Z - P_y \|_2
\end{equation}
We hypothesize that the MCI and AD cohorts will show significantly higher mean BAG and PAE values, quantifying their deviation from the learned normative manifold. We use two-sample t-tests to confirm the statistical significance of these differences.

\section{Results}
\subsection{Performance Comparison}

We present the results of the proposed EVA-Net and all baseline models on the primary task of normative age regression, evaluated on the 10\% held-out healthy test sets from our 10-fold cross-validation.

Table~\ref{tab:performance_comparison} summarizes the performance according to the MAE, RMSE, and $R^2$ metrics. Our proposed EVA-Net achieves state-of-the-art performance, outperforming all competitive baselines on all three metrics. Notably, EVA-Net achieves an MAE of 2.645, a 13.4\% improvement over the strongest baseline, MBN (MAE of 3.053). Similarly, EVA-Net obtains the lowest RMSE (3.327) and the highest $R^2$ (0.939), indicating that its predictions are not only accurate but also explain a larger proportion of the variance in chronological age. 

\subsection{Ablation Study}
\label{sec:ablation}
\begin{table}[t]
    \centering
    \caption{Ablation study of EVA-Net's core components. Results are averaged across the 10-fold cross-validation on the healthy test set. The full model's performance is compared against variants lacking the Efficient backbone (w/o E), the Variational bottleneck (w/o V), and the Alignment loss (w/o A).}
    \label{tab:ablation}
    \resizebox{\columnwidth}{!}{%
    \begin{tabular}{lccc}
        \toprule
        \textbf{Model Variant} & \textbf{MAE $\downarrow$} & \textbf{RMSE $\downarrow$} & \textbf{$R^2$ $\uparrow$} \\
        \midrule
        \textbf{EVA-Net (Full Model)} & \textbf{2.645} & \textbf{3.327} & \textbf{0.939} \\
        \midrule
        EVA-Net w/o E (Efficient) & 5.138 & 6.350 & 0.881 \\
        EVA-Net w/o V (Variational) & 5.512 & 7.104 & 0.866 \\
        EVA-Net w/o A (Alignment) & 5.903 & 7.822 & 0.849 \\
        \bottomrule
    \end{tabular}%
    }
\end{table}

To validate the contribution of each key component in our proposed EVA-Net framework, we conducted a comprehensive ablation study. We evaluated three distinct variants of our model on the 10-fold cross-validation dataset:
\begin{itemize}
    \item \textbf{EVA-Net w/o E (Efficient):} In this variant, we removed the (E)fficient ProbSparse attention backbone. We replaced the Informer-based encoder with a standard GRU-based recurrent neural network to model the temporal sequence, while keeping the VIB and Alignment components.
    \item \textbf{EVA-Net w/o V (Variational):} We removed the (V)ariational Information Bottleneck. The hidden representation $H_i$ from the efficient encoder was passed directly and deterministically to the prediction head $f_{\psi}$ and the prototype alignment $L_{align}$. The $L_{IB}$ loss term was removed from the training objective.
    \item \textbf{EVA-Net w/o A (Alignment):} We removed the (A)lignment component entirely. The continuous prototype network $P_{\theta}$ was removed, and the $L_{align}$ loss was omitted. The model was trained only to optimize the prediction and VIB losses ($L_{pred} + \beta L_{IB}$).
\end{itemize}

The results, presented in Table~\ref{tab:ablation}, demonstrate the critical and synergistic role of all three components. The full EVA-Net (MAE 2.645) significantly outperforms all ablated models. The removal of any single component, Efficient, Variational, or Alignment, causes a catastrophic drop in performance. Notably, removing the Alignment module (w/o A) caused the most significant performance degradation, highlighting the importance of the prototype-guided latent space structure for learning a generalizable normative model.

\subsection {Sensitivity Study}
\begin{figure}
    \centering
    \includegraphics[width=0.96\linewidth]{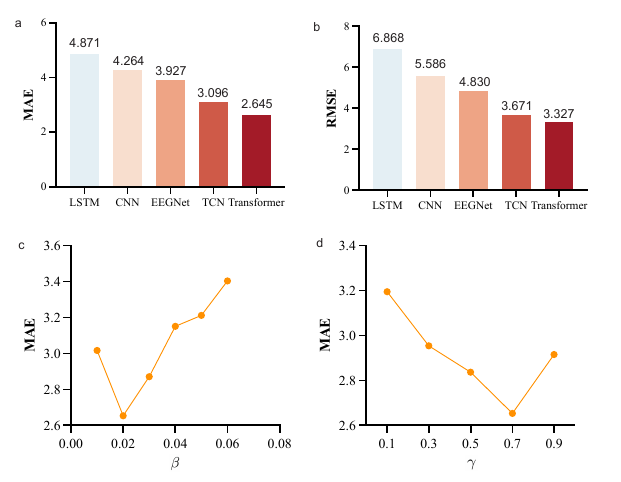}
    \caption{Sensitivity analysis of EVA-Net. (a, b) MAE and RMSE comparison across encoder backbones. (c) VIB loss weight $\beta$ sensitivity within \{0.01, 0.02, 0.03, 0.04, 0.05, 0.06\}. (d) Prototype alignment loss weight $\gamma$ sensitivity within \{0.1, 0.3, 0.5, 0.7, 0.9\}, optimal at $\gamma=0.7$.}
    \label{fig:sensitivity}
\end{figure}
To validate our architectural choices and hyperparameter settings, we conducted a sensitivity analysis, with results presented in Figure~\ref{fig:sensitivity}.

First, we evaluated the impact of our Transformer-based backbone by comparing its performance against four other widely-used time-series encoders: LSTM, a standard CNN, EEGNet, and TCN. As shown in Figure~\ref{fig:sensitivity}a and \ref{fig:sensitivity}b, the Transformer architecture achieves the best performance by a significant margin.It yields the lowest MAE of 2.645 and the lowest RMSE, substantially outperforming the next-best model, TCN, as well as EEGNet, CNN, and LSTM. This result confirms that the long-range dependency modeling of the Transformer is highly effective for this task and justifies its selection as our primary encoder.

Second, we analyzed the model's sensitivity to the two key loss-weighting hyperparameters in our composite loss function: $\beta$ (for $L_{IB}$) and $\gamma$ (for $L_{align}$). For the $\beta$ sensitivity test (Figure~\ref{fig:sensitivity}c), we evaluated values within \{0.01, 0.02, 0.03, 0.04, 0.05, 0.06\}. The model achieves its best performance with a small $\beta$ (e.g., $\beta=0.01$), with the MAE degrading as the VIB compression penalty increases, which justifies our selection of a small $\beta$ value. For the $\gamma$ sensitivity test (Figure~\ref{fig:sensitivity}d), we evaluated values within \{0.1, 0.3, 0.5, 0.7, 0.9\}. The plot shows a clear U-shaped curve: the MAE decreases from 3.2 at $\gamma=0.1$, reaches its global minimum of 2.645 at $\gamma=0.7$, and then rises again at $\gamma=0.9$. This result demonstrates that the prototype alignment is critical for performance, but must be balanced against the primary prediction task, validating our final choice of $\gamma=0.7$.

\subsection{Anomaly Detection on Pathological Cohorts}
\label{sec:anomaly_detection}

\begin{figure}
    \centering
    \includegraphics[width=0.96\linewidth]{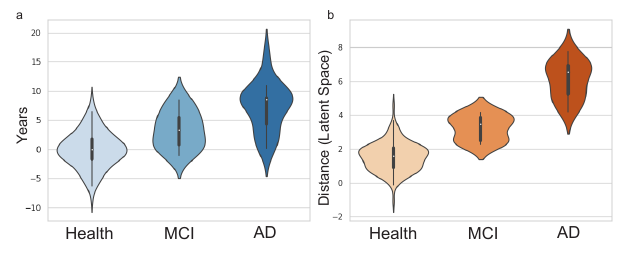}
    \caption{Violin plots illustrating the distribution of anomaly scores across cohorts. (a) Brain-Age Gap (BAG): While the healthy control group (N=130) is centered around zero, the MCI (N=12) and AD (N=15) groups show progressively larger positive gaps, indicating accelerated brain aging. (b) Prototype Alignment Error (PAE): A similar trend is observed for PAE, where pathological groups exhibit significantly larger distances from the normative manifold. The violin shapes visualize the full density of the data, highlighting the clear separation between healthy and pathological distributions.}
    \label{fig:anomaly_results}
\end{figure}

Having established EVA-Net's state-of-the-art regression performance on the healthy cohort, we tested our core hypothesis: that the learned normative manifold can serve as an effective anomaly detector for pathological aging. We froze the trained model and applied it to the unseen pathological cohort (12 MCI and 15 AD subjects).

As defined in our Evaluation Metrics, we computed both the Brain-Age Gap (BAG) and the Prototype Alignment Error (PAE) for all subjects. The distributions of these anomaly scores are visualized in Figure~\ref{fig:anomaly_results}. As shown, the healthy test subjects exhibited a minimal BAG ($0.2 \pm 3.3$ years) and a low baseline PAE ($1.5 \pm 0.8$), confirming that the model accurately maps healthy individuals to the normative manifold. In contrast, the pathological groups demonstrated progressive distributional shifts away from the healthy norm. The MCI group showed a significantly elevated mean BAG of $3.7 \pm 3.1$ years and a PAE of $3.1 \pm 1.2$. The AD group exhibited the most severe deviation, with a mean BAG reaching $7.4 \pm 4.5$ years and a PAE of $5.6 \pm 1.7$. These results confirm a clear trend ($\text{AD} > \text{MCI} > \text{Healthy}$) where disease severity strongly correlates with the magnitude of deviation from the learned healthy prototype, validating PAE as a sensitive, interpretable marker for pathology.

\section{Conclusion}
\label{sec:conclusion}

In this work, we presented EVA-Net, a novel framework for normative brain-age modeling and anomaly detection from imperfect, healthy-only EEG data. By integrating an efficient ProbSparse attention backbone, a Variational Information Bottleneck, and an age-conditioned continuous prototype network, EVA-Net successfully learns a robust, compressed, and geometrically interpretable manifold of healthy aging. Our extensive experiments demonstrate that EVA-Net not only achieves state-of-the-art regression accuracy on healthy subjects (MAE of 2.645 years) but also serves as a powerful anomaly detector. The proposed Prototype Alignment Error (PAE) effectively quantified pathological deviations in unseen MCI and AD cohorts, revealing a clear progression of disease severity. This study establishes a new, interpretable paradigm for leveraging deep learning on widely available clinical EEG data to screen for neurodegenerative conditions, paving the way for low-cost, accessible brain health monitoring.

\section*{Acknowledgment}

This work was supported by the National Natural Science Foundation of China (Grant No. 81702469) and the Natural Science Foundation of Shandong Province (Grant No. ZR2023MH321).

This work involved human subjects in its research. Approval of all ethical and experimental procedures and protocols was granted by the Ethics Committee of Qilu Hospital of Shandong University under Protocol No. KYLL-202406 (YJ) -010.

\bibliographystyle{unsrt}  
\bibliography{references}

\end{document}